\documentclass[letterpaper, 10 pt, conference]{ieeeconf}  

\IEEEoverridecommandlockouts                              

\usepackage{amsmath,amssymb,amsfonts}
\usepackage{algorithm}
\usepackage{graphicx}
\usepackage{xcolor}

\usepackage{footnote}
\makesavenoteenv{tabular}
\usepackage{algpseudocode}
\usepackage{multicol}
\usepackage{float}
\usepackage{epstopdf}	
\usepackage{longtable}
\usepackage{comment}
\usepackage{xspace}

\usepackage{tikz}
\usepackage{transparent}

\usepackage{tabularx,ragged2e,booktabs}

\usepackage{mathtools}

\usepackage{tablefootnote} 

\usepackage{multicol} 
 
\usepackage{nomencl} 
\makenomenclature
\renewcommand*\nompreamble{\begin{multicols}{2}}
\renewcommand*\nompostamble{\end{multicols}}

\makeatletter
\DeclareRobustCommand\onedot{\futurelet\@let@token\@onedot}
\def\@onedot{\ifx\@let@token.\else.\null\fi\xspace}

\def\eg{\emph{e.g}\onedot}

\def\etal{\emph{et al}\onedot}
\makeatother

\newcommand{\guy}[1]{\textcolor{red}{guy: #1}}

\begin{document}

\title{\LARGE \bf Hypergraph-Transformer (HGT) for Interaction Event Prediction \\
in Laparoscopic and Robotic Surgery 

\thanks{
}

}

\author{Lianhao Yin$^{1,2}$, Yutong Ban$^{3}$, Jennifer Eckhoff$^{1,5}$, Ozanan Meireles$^{1,4}$, Daniela Rus$^{2}$, Guy Rosman$^{1,4}$ 
\thanks{ $^{1}$ Surgical Artificial Intelligence Laboratory, Massachusetts General Hospital, MA, US, {\tt\small lyin5@mgh.harvard.edu ,jennifer.eckhoff@uk-koeln.de, rosman@csail.mit.edu}.}
\thanks{$^{2}$ Computer Science and Artificial Intelligence Laboratory, MIT, MA, US
        {\tt\small lianhao,rus@csail.mit.edu}.}
\thanks{ $^{3}$ UM-SJTU Joint Institute, Shanghai Jiao Tong University, Shanghai, China {\tt\small yban@sjtu.edu.cn} }
\thanks{ $^{4}$ Surgical Artificial Intelligence Laboratory, Department of Surgery, Duke, NC, US {\tt\small guy.rosman,ozanan.meireles@duke.edu} }
\thanks{ $^{5}$ University Hospital Cologne, Department of Visceral, General, Thoracic and Transplant Surgery, Cologne Germany {\tt\small jennifer.eckhoff@uk-koeln.de} }
\thanks{Acknowledgement: The authors receive partial research support from the Department of Surgery
at Massachusetts General Hospital as well as CRICO. The research work was done when Yutong Ban was at MIT and MGH, and Jennifer Eckhoff were affiliated with MGH}
\thanks{Copy right: Personal use of this material is permitted.  Permission from IEEE must be obtained for all other uses, in any current or future media, including reprinting/republishing this material for advertising or promotional purposes, creating new collective works, for resale or redistribution to servers or lists, or reuse of any copyrighted component of this work in other works.}}

\maketitle
\thispagestyle{empty}
\pagestyle{empty}

\begin{abstract}
Understanding and anticipating events and actions is critical for intraoperative assistance and decision-making during minimally invasive surgery. 
We propose a predictive neural network that is capable of understanding and predicting critical interaction aspects of surgical workflow based on endoscopic, intracorporeal video data, while flexibly leveraging surgical knowledge graphs. The approach incorporates a hypergraph-transformer (HGT) structure that encodes expert knowledge into the network design and predicts the hidden embedding of the graph. We verify our approach on established surgical datasets and applications, including the prediction of action-triplets, and the achievement of the Critical View of Safety (CVS), which is a critical safety measure. Moreover, we address specific, safety-related forecasts of surgical processes, such as predicting the clipping of the cystic duct or artery without prior achievement of the CVS. Our results demonstrate improvement in prediction of interactive event when incorporating with our approach compared to unstructured alternatives. 
\end{abstract}

\section{Introduction} \label{se:introduction}
Imagine a surgeon performing minimally invasive robotic surgery (MIRS), while AI serves as a copilot, anticipating adverse events and advising the surgeon on subsequent actions to take. This could look like AI recommending a different course of action in anticipation of an imminent complication or issuing a warning if the surgeon's actions deviate from the recommended course of action. For example, as the key anatomic structures must be clipped before being cut, the predictive system could issue a warning if the structures are about to be cut before they are clipped. Besides reducing surgeons' cognitive workload, such a prediction system would prevent adverse events and improve surgical safety.

For AI to assist and benefit surgeons intraoperatively, surgical robotic and assistive systems need to leverage the endoscopic video data to predict granular events, interactions, and dependencies in the future related to surgical workflow and decision-making. Similar to other safety-critical domains, such as driving \cite{Ivanovic2020-ns,Balachandran2023-lf,Rudenko2020-cp} or socially-aware robotics \cite{Truong2018-mp, Mavrogiannis2023-cp,Fridovich-Keil2020-qw}, and industrial robotics \cite{Kanazawa_undated-cj,cai2023prediction,li2024safe}, being able to predict human actions and their future consequences is key to assistive systems. In surgery, the premise is that accurate prediction can lead to timely alerts for the surgeon, enabling corrective actions before complications occur.

\begin{figure}[t!] 
	\begin{center}
        \includegraphics[width=0.40\textwidth]{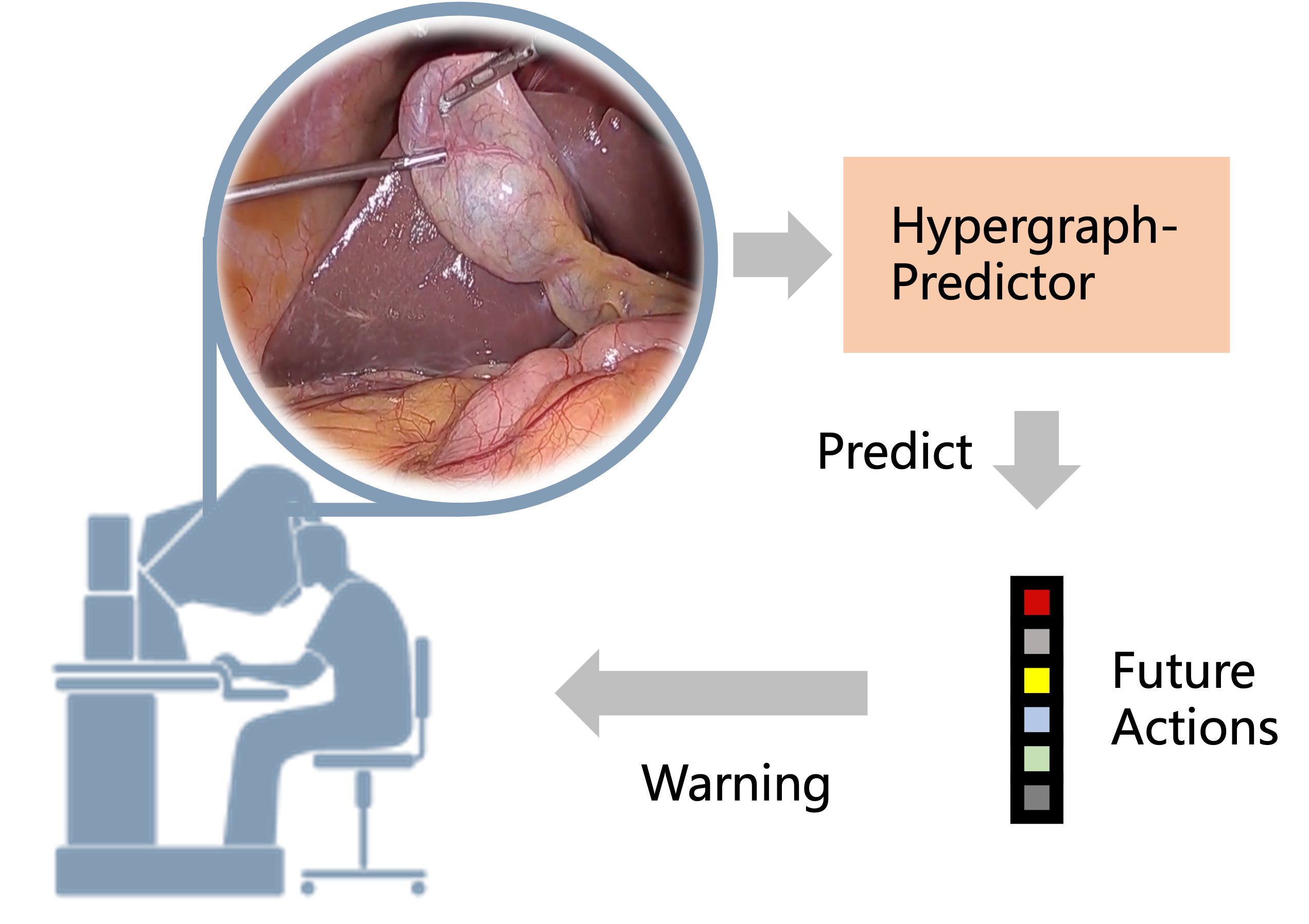}
     	\caption{We propose an event prediction framework during surgery based on a Hypergraph Transformer (HGT). The model makes predictions of action-triplets, and surgical safety measures based on recent past video frames.}
   	\label{fig:teaser}
	\end{center}
\end{figure} 

In this paper, we develop a general framework for predicting safety-related events in endoscopic surgery. We compute sample sequences of labels that pertain to surgical events, conditioned on the previous video frames, at the prediction time point. 
We focus on laparoscopic cholecystectomy \cite{cameron1991laparoscopic, Soper1991-bb} as a benchmark procedure in surgical data science, based on its standardized workflow and high volume(approximately 750,000 US cases annually) \cite{Vollmer2007-ri}.

We examine the proposed approach in predicting three types of interaction events, based on recent video frames. The first type of prediction focuses on action-triplets. Surgical action-triplets refer to the interactions of three components: the tool, the performed action, and the target object (\eg tissue or anatomic structures). For example, tools include scissors, graspers or clip-appliers, the performed actions can be cutting, clipping or grasping, and the target can be any anatomic structure or tissue such as the gallbladder or cystic duct. In laparoscopic cholecystectomy, there are 100 viable combinations of action-triplets \cite{Nwoye2020-qm}. Due to the high variability of visual content and the inherent limitation of monocular vision, automated detection of action-triplets is challenging~\cite{Nwoye2023-ci}. The second type of event is the achievement of Critical View of Safety (CVS), introduced by Strasberg \cite{Strasberg1995-jy}. The CVS is a widely established surgical safety measure that precedes irreversible steps (i.e. clipping and division of the cystic duct or artery before removal of the gallbladder) to prevent biliary injuries \cite{Strasberg2010-eq}. The third event type we predict involves safety-critical irreversible actions occurring before necessary prerequisites and safety checks have been completed. An example for this would be the clip-applier clipping a ductal structure before the CVS is achieved. This prediction of such events is key to preventing surgical complications, such as irreversible damage to critical anatomic structures (e.g. the common bile duct).

In this paper, we address these three risk-related predictive tasks in MIRS using our framework. 
Specifically, the contributions of our work are: 
\begin{itemize}
    \item Development of a general framework using hypergraph-transformer encoder-decoders for interaction event prediction in minimally invasive robotic surgery. 
    \item Introduction of several novel clinically relevant prediction tasks, such as detection and prediction of action-triplets, the CVS, and compositions of the two in the context of surgical safety, based on publicly available datasets. 
    \item Demonstration of state-of-the-art results for predicting surgical events such as action-triplets compared to detection, offering strong implications for AI-based surgical safety systems in MIRS.
\end{itemize}

\section{Related works} \label{se: related work}

\subsection{Methods for Detecting Interaction Events in MIRS}
This paper focuses on interactions in laparoscopic cholecystectomy. The vast majority of current works in surgical data science focuses on the detection of interaction events in current frames, rather than prediction in future frames. For example, the Tripnet architecture~\cite{Nwoye2020-qm} achieved an average precision of 29\% taken over 100 action-triplets detection during laparoscopic cholecystectomy. More broadly, detecting tool-tissue interactions has been studied using graph convolutions \cite{Ulutan2020-mf, Maraghi2021-tb, Rupprecht2016-si, Mallya2016-ip, Ban2023-ey}, and attention mechanisms \cite{Nwoye2020-qm, Ban2023-ey}. In surgery, there is high deformation and visual variability of tissue, thus interaction events are complicated to detect. With limited temporal context, it is even harder to predict interaction events in the future, further limiting the accuracy expected in action-triplet prediction. 

\subsection{Methods for Prediction of Future Interaction-Events}
Since surgical interactions are recorded as videos, they can be treated as sequential image data. The prediction of future interaction events is equivalent to predicting the event in the future using the sequential image data from past. 

\subsubsection{Sequential data prediction}
Sequential data prediction has often been incorporated in autoregressive methods such as LSTM \cite{Hochreiter1997-gm}. LSTMs have been applied to predict the trajectories of different systems such as trajectories of vehicles in traffic\cite{Salzmann2020-pg}, pedestrians' path~\cite{Alahi2016-zq, Mohamed2020-hu}, and players' actions in sports\cite{Li2020-jp}. These methods' autoregressive structure limits long-term prediction due to gradient vanishing and exploration, and they have been more recently replaced by transformer-based approaches \cite{Vaswani2017-bn, Ngiam2022-ny}. Transformer approaches using attention mechanisms have also been used in surgical video analysis \cite{sharma2023rendezvous}. To date, using a single attention-based framework to capture the vast variety of raw inputs and broad semantics pertaining to surgery has proven difficult\cite{sharma2023rendezvous}.

\subsubsection{Generative adversarial networks} Generative adversarial networks (GAN)\cite{Goodfellow_undated-qk, Creswell2018-ca} have been widely used for video \cite{Liang2017-sd} and sequence prediction \cite{Ban2022-gl}, especially human motion prediction for social navigation and autonomous driving \cite{Gupta2018-go, Salzmann2020-pg, Huang2020-pr}. In a surgical context, Ban \etal\cite{Ban2022-gl} proposed a GAN-based approach for surgical workflow prediction, demonstrating its effectiveness in predicting the progress of laparoscopic cholecystectomy. However, the approach was specific to surgical phase prediction, and not generalized to prediction of multiple facets of the surgical workflow or more granular aspects pertaining to the course of a surgical procedure. 

\subsubsection{Graph neural networks} Graph neural networks (GNN) \cite{Scarselli2009-tl} model complex interactions by representing objects as nodes and their relations as edges. This approach has been widely used for prediction of trajectories~\cite{Salzmann2020-pg}, as well as surgical workflow analysis \cite{Ban2023-ey,Murali2024-at}. The branches of GNNs can make causal predictions~\cite{Xu2020-of}, and were therefore widely used to encode complex relations. A special form of GNNs is the hypergraph architecture \cite{Feng2019-wz, bretto:hal-01024351, Ban2023-ey}. So far hypergraphs have not been used for prediction of interaction events in surgery and more specifically MIRS. One reason for this is the fast-paced dynamics inherent to surgery, requiring constantly evolving propagation of the graph's relations.

Our paper incorporates hypergraphs with transformer encoders to predict complex surgical interactions through relation propagation. The main motivation is to leverage surgical knowledge in the graph's structure, resulting in a sparse and yet informative representation for critical relations that affect safety in MIRS. We use transformers to predict the hidden state of the graph along the time horizon in order to enable the predictions of the surgical concepts. 
\begin{figure}[t]
    \centering
    \includegraphics[width=0.45\textwidth]{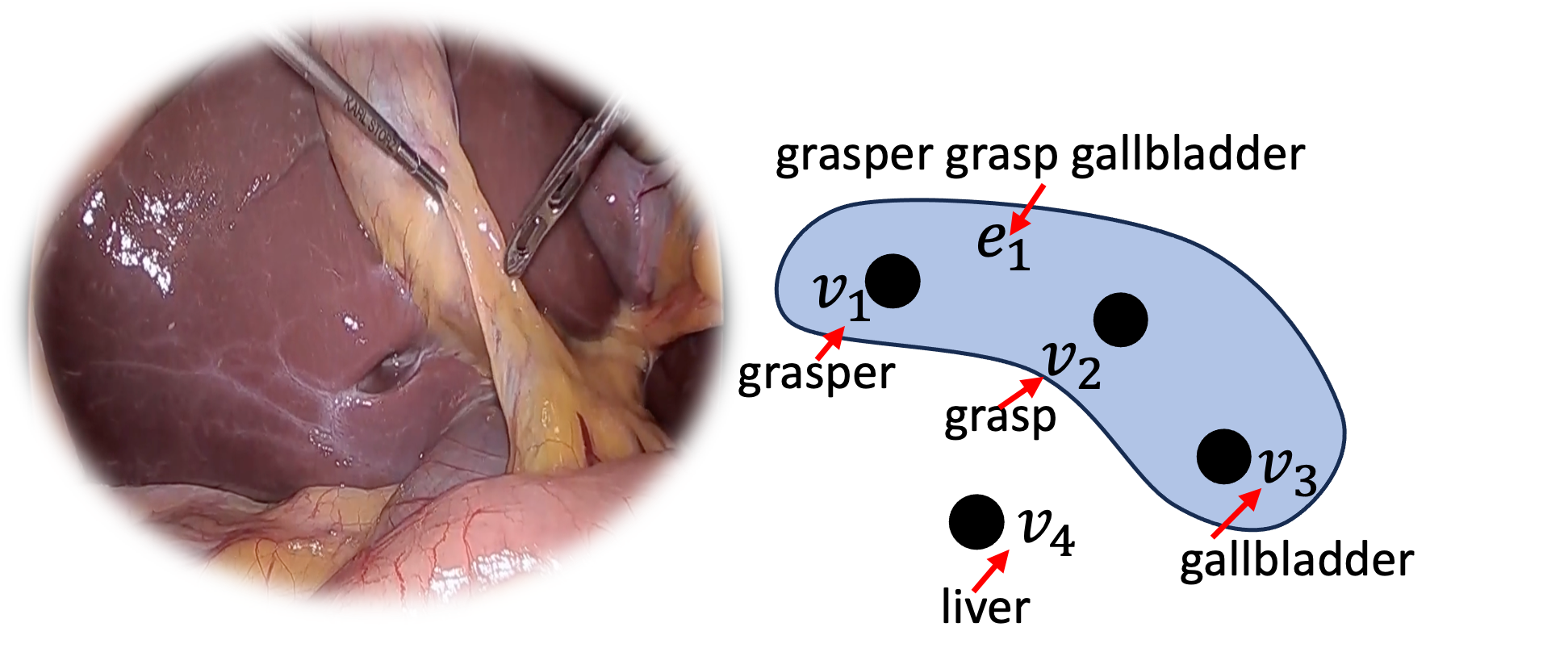}
    \caption{Example of edge and nodes. The subgraph on the right captures the relevant concepts and a hyperedge for the video frame on the right \cite{Roy2020-is}.}
    \label{fig:example of edge and nodes}
\end{figure}
\begin{figure*}[th] 
    \begin{center}
        \includegraphics[width=1.0\textwidth]{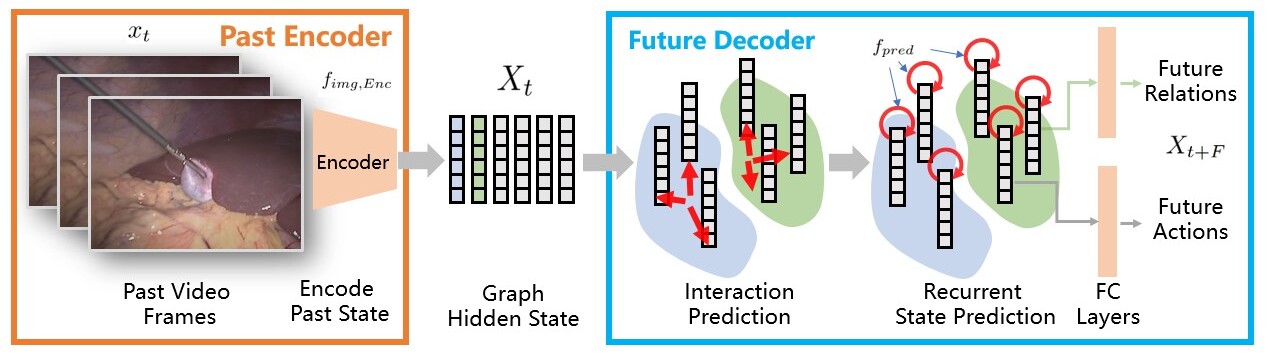}
        \caption{The architecture of the proposed Hypergraph-transformer. It is to make predictions of action sequences by using past video frames. The encoder encodes the past video frames and the decoder makes predictions. A pre-trained DINO was used as the visual backbone. The encoder and decoder use hypergraph message passing for every graph element separately. }
        \label{fig: neural network structure}
    \end{center}
\end{figure*}
\section{Methods} \label{se: model}
\subsection{Notations}
The neural network outputs are referred to as "detections" for events or labels at the current time, while the term "predictions" is reserved for events or labels in the future. 
The variable as $x_{t} \in \mathbf{R}^{W,L,3}$ represent input images where the subscript $t$ is the time index of that frame, and the images have a width of $W$, a height of $L$, and a channel dimension of 3. The hidden state of $i$th edge $e$ is represented by $X_{t, e_i}$ at $t$ time index. The system's output is represented by $y_{t}$ at $t$ time index. Action-triplets are denoted as $\langle\textit{tool, action, target}\rangle$. Neural networks are denoted by $f$.

\subsection{Problem Statement}
We define the concept-structured prediction problem in a surgical context as follows. Given a recent sequence of laparoscopic video frames $\{x_{t-P+1},...,x_{t}\}$ from $P$ seconds before the current time $t$, we predict how different entities change over time in the future of the surgical video.

The presence or visibility of surgical concepts at time $t$, is annotated using binary labels $y_{t}$. For example, the presence of a clip-applier at time $t$ is annotated using $y_{t, \langle\textit{clip-applier}\rangle}$; the clip-applier clipping the cystic duct at time $t$ is annotated using $y_{t, \langle\textit{clip-applier,clips,cystic-duct}\rangle}$.

The task is to predict the future labels based on the current image inputs. Formally, given a sequence of inputs $x_{t-P+1},\ldots,x_t$, predict the labels of $\hat{y}_t,\ldots,\hat{y}_{t+F}$, where $F$ is the prediction horizon (short for hp in the tables). This task is essential for the development of computer-assisted monitoring and assistive systems targeting time-critical notification and guidance of surgeons.

\subsection{Neural Network Architecture}
\subsubsection{Graph Design}
The proposed graph structure is based on surgical expert knowledge~\cite{Ban2023-ey}. Specifically, the nodes $V_t = \{v_{t,1},...,v_{t, i},...\}$ at each time point represent surgical concepts, including tools (such as \textit{clip-applier} or \textit{irrigator}), tissues or organs (\eg the \textit{liver}, \textit{gallbladder}, or the \textit{cystic duct}); the edges $E_t=\{e_{t,1},...,e_{t,j},...\}$ at any given time point are the interactions between the nodes, i.e. between tools and tissue, (\eg \textit{cut} and \textit{grasp}). For example, the hyperedge, or relation, $\langle \textit{clip-applier,clips,cystic-duct}\rangle$ captures the interaction between the 3 concepts, forming an action-triplet \cite{Nwoye2020-qm}. To illustrate, figure~\ref{fig:example of edge and nodes} shows a surgical video frame, in which the gallbladder is being grasped by the grasper, while the liver is present in the background. The corresponding subgraph correctly shows the nodes representing these concepts $\langle\textit{grasper, gallbladder, liver}\rangle$ as well as the edge representing the interaction between them as $\langle\textit{grasper, grasp gallbladder}\rangle$. The connection of the nodes and edges for each task is based on annotations by surgical experts, see Sec.~\ref{se: results} (Fig.~\ref{fig:subgraphs}) for details.

As shown in Fig.~\ref{fig: neural network structure}, the neural network comprises an encoder for the past video frames and a decoder to predict embeddings. The encoder and decoder both use a hypergraph-transformer structure. We used a pre-trained vision transformer DINO \cite{Dosovitskiy2020-dp,Caron2021-rb} as the image encoder $f_{img,Enc}$ backbone (Eq.~\ref{eq: img_encoder}).
The pretrained DINO has shown to be performant, while keeping interpretable saliency maps \cite{Dosovitskiy2020-dp,Caron2021-rb}. It was pre-trained using ImageNet data. The images in the dataset in this paper are a subset of the image from a diversity of object perspectives, although there are objects such as tissues not included in ImageNet data. Considering this, the last two layers of the backbone were unfrozen, and we added another layer for fine-tuning.
\begin{equation} \label{eq: img_encoder}
    \begin{aligned}
        X_{t,e_i} = f_{img,Enc,e}(x_t) \\
        X_{t,v_i} = f_{img,Enc,v}(x_t)
    \end{aligned}
\end{equation}

The edge feature of $i$th edge $e_i$ is represented by $X_{t, e}$ and the node feature of $j$th node $v_j$ is represented by $X_{t, v}$ at time $t$. Specifically, the $i$th edge $e_i$ is connected with a set of nodes $v_{j},...,v_{j+S}$. The selection of nodes to be connected is determined by the graph annotations based on the surgical task. The $j$th node is connected with a set of edges $e_{i},...,e_{i+N}$ and which edge to be connected is according to the annotation as well. The hidden state of each edge and node is updated recurrently according to the model's temporal information.


At each message passing, The embedding of edge (Eq.~\ref{eq: edge features}) and the embedding of node (Eq.~\ref{eq: node features}) are updated recursively. This enables the information of relations and object detection to be passed to their hidden representations.
\begin{equation} \label{eq: edge features}
    \begin{aligned}
        X_{t, e_i} = f_{E,Enc}(X_{t, v_{j}},...,X_{t, v_{j+S}}, X_{t, e_i})
    \end{aligned}
\end{equation}
\begin{equation} \label{eq: node features}
    \begin{aligned}
        X_{t, v_{j}} = f_{V,Enc}(X_{t, e_i},...,X_{t, e_{i+N}}, X_{t, v_j})  
    \end{aligned}
\end{equation}
The encoder of edge $f_{E,Enc}$ and node $f_{V,Enc}$ are both with 2 linear layers and ReLU activation. We adopted batch-norm in each layer to stabilize and speed up the training. In both encoder and decoder, we compute the overall structure sequentially, advancing in time and combining message passing and past frame encoding.

\subsubsection{Prediction with transformer }
We incorporate the hypergraph-transformer to predict the recurrent states of edge $X_{t+1,e_i}$ and node $X_{t+1,v_i}$ at the next time points. Unlike node-centric approaches (\eg\cite{Velickovic2017-xf,Ngiam2021-hn}), which use transformers to cross attention among nodes, the transformer in this paper predicts embedding of each node and edge seen in Fig.~\ref{fig: neural network structure}. The main reason behind is that the hypergraph structure already encodes the cross-attention through different nodes in the graph design. 
Specifically, the transformer is used in the recurrent state on both edge and node $X_{e_i}$ and $X_{v_i}$. 
From the feature $X_t$ at current time $t$ to predict the feature at $t+F$, one can model the hidden state propagation by 
\begin{equation} \label{eq:embedding prediction}
    \begin{aligned}
        X_{t+F, e_i} = (f_{pred} \circ f_{pred} ... \circ f_{pred})X_{t, e_i} \\
        X_{t+F, v_i} = (f_{pred} \circ f_{pred} ... \circ f_{pred})X_{t, v_i} \\
    \end{aligned}
\end{equation}
The predictor $f_{pred}$ is the transformer neural network that interweaves message passing and temporal state updates for the concepts and relations in the graph. We use a multi-head transformer with 2 heads and 2 layers as predictor $f_{pred}$. 

A projector $f_{Proj}$ using linear layers decodes the hidden state and maps to the labels in Eq.~\ref{eq: projector}.
\begin{equation} \label{eq: projector}
    \begin{aligned}
\hat{y}_{t+F, v,e} = f_{Proj}(X_{t+F, v}, X_{t+F,e})
    \end{aligned}
\end{equation}
We use binary cross-entropy (BCE) loss $l_{loss}$ to measure the difference between $y_{t+F}$ and $\hat{y}_{t+F}$, i.e., $l_{loss} (\hat{y}_{t+F}, y_{t+F})$.

\begin{algorithm} 
\caption{Algorithm for the proposed approach }\label{alg: proposed method}
\begin{algorithmic}[1]
\For{each epoch}
\State Resample/Balance data
\For{each video clip}
    \State $/*$ Encoding:$*/$
    \For{Each time point in video}
    \State Get image embeddings Eq.~\ref{eq: img_encoder}
    \State $/*$ Message passing $*/$
    \State Update the edge feature Eq.~\ref{eq: edge features}
    \State Update the node feature Eq.~\ref{eq: node features}
    \EndFor
    \State $/*$Decoding:$*/$
    \For{Each time point in prediction}
    \State $/*$ Message passing $*/$
    \State Update the edge feature Eq.~\ref{eq: edge features}
    \State Update the node feature Eq.~\ref{eq: node features}
    \State Predict the embeddings Eq.~\ref{eq:embedding prediction}
    \State Predict the output with projector Eq.~\ref{eq: projector}
    \EndFor
\EndFor
\State Backpropagation $l_{loss} (\hat{y}_{t+F}, y_{t+F})$.
\EndFor
\end{algorithmic}
\end{algorithm}
\subsection{Training}
Due to the sparsity of positive labeling in the dataset, this paper adopted importance sampling, which is similar to focal loss \cite{Lin2020-of}, during training to balance the training labels to improve the accuracy of the predictions.
The training algorithm of the proposed method is summarized in Algorithm \ref{alg: proposed method}. 

As explained, the hidden state of the edge depends on the node's features. The authors defined two training phases. The first phase pre-trained the node features first to decompose the dependency of edge and node features. The second phase trained the edge and node features together to predict the events of interest. Intuitively, the first phase weights more on learning the spatial and contextual information of objects and events from the video inputs, and the later phase weights more on learning the critical events such as action-triplets.  

\begin{figure*}[t]
    \begin{minipage}[t]{.24\textwidth}
    \centering
    \vspace*{2mm}
    \includegraphics[width=0.8\textwidth]{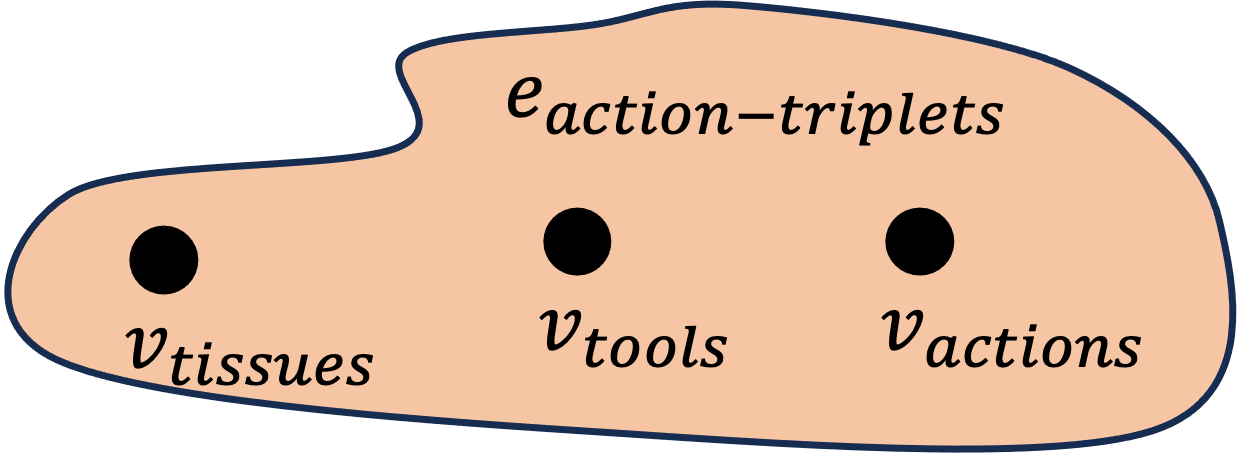}

    (a)
    \label{fig:node_edge for action triplet}
    \end{minipage}
    \hfill
    \begin{minipage}[t]{.24\textwidth}
    \centering
    \vspace*{2mm}
     \includegraphics[width=\textwidth]{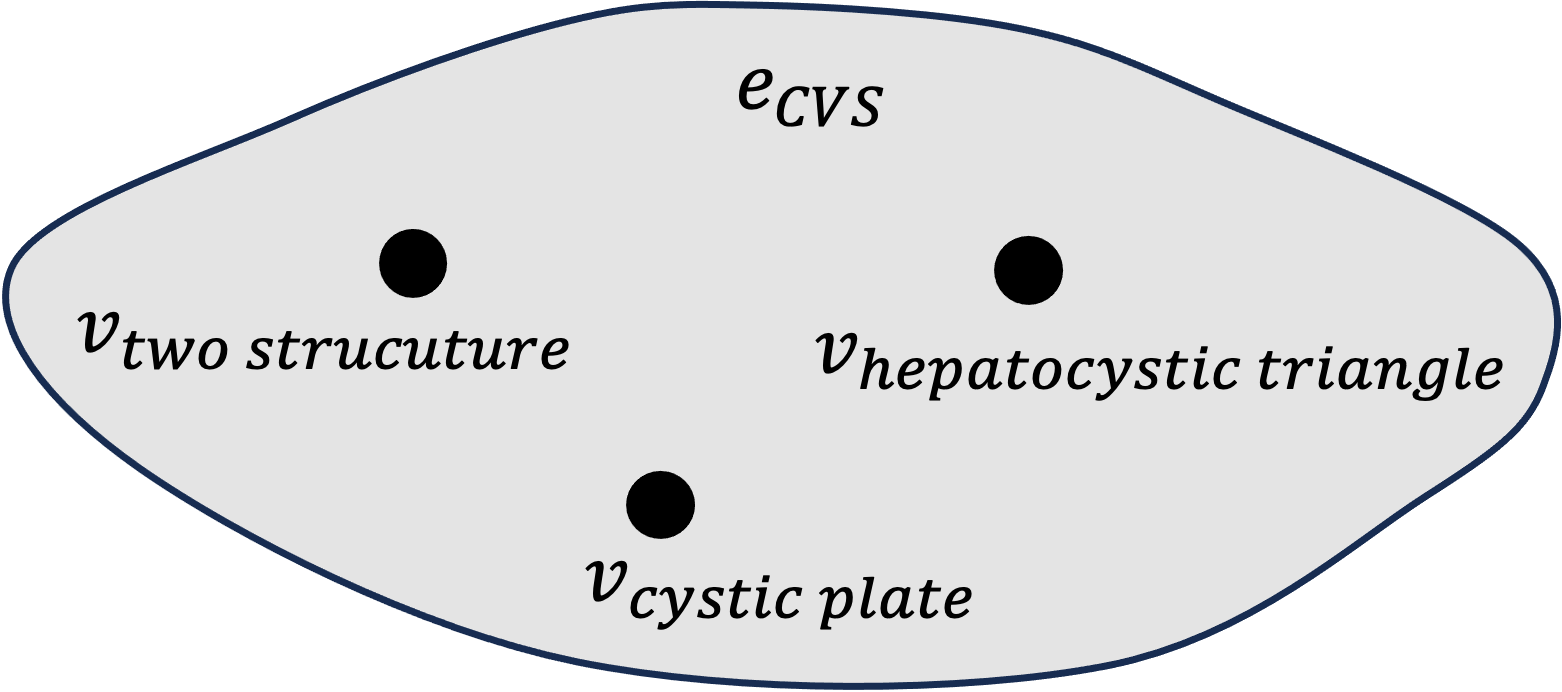}

    (b)
    \label{fig:node_edge for CVS}
    \end{minipage}  
    \hfill
    \begin{minipage}[t]{.24\textwidth}
    \centering
    \vspace*{2mm}
    \includegraphics[width=\textwidth]{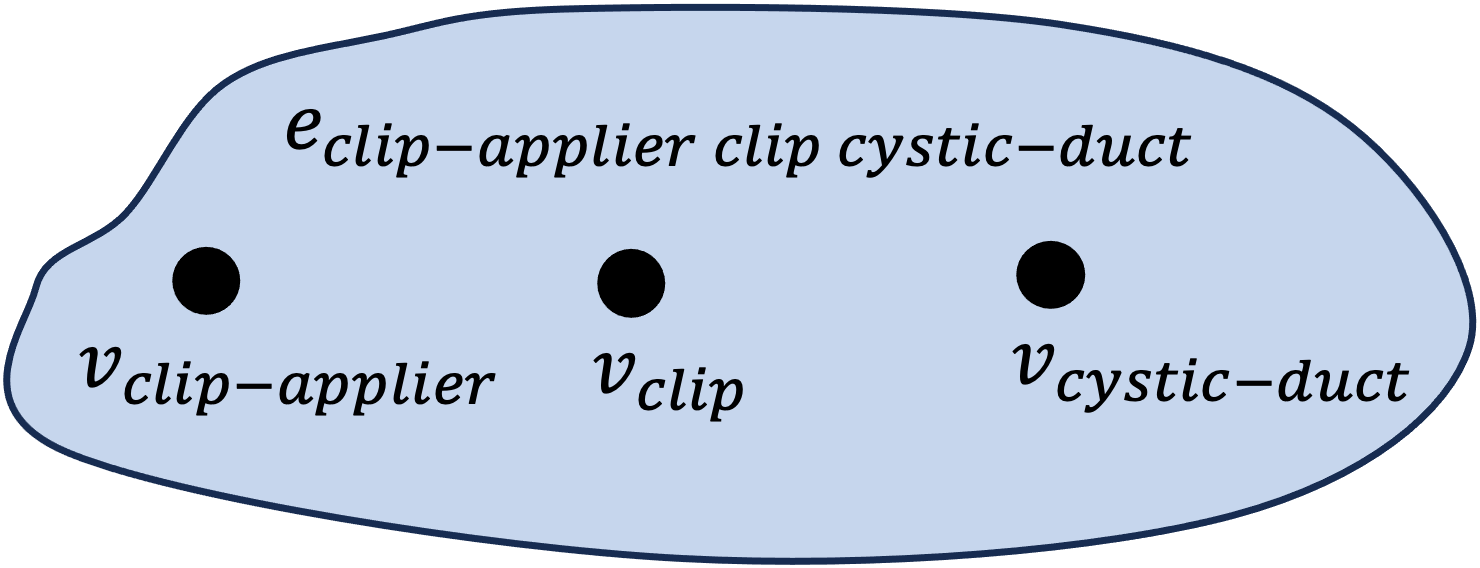}

    (c)
    \label{fig:node_edge for clipping}
    \end{minipage}  
    \hfill
    \begin{minipage}[t]{.24\textwidth}
    \centering
    \vspace*{2mm}
    \includegraphics[width=0.85\textwidth]{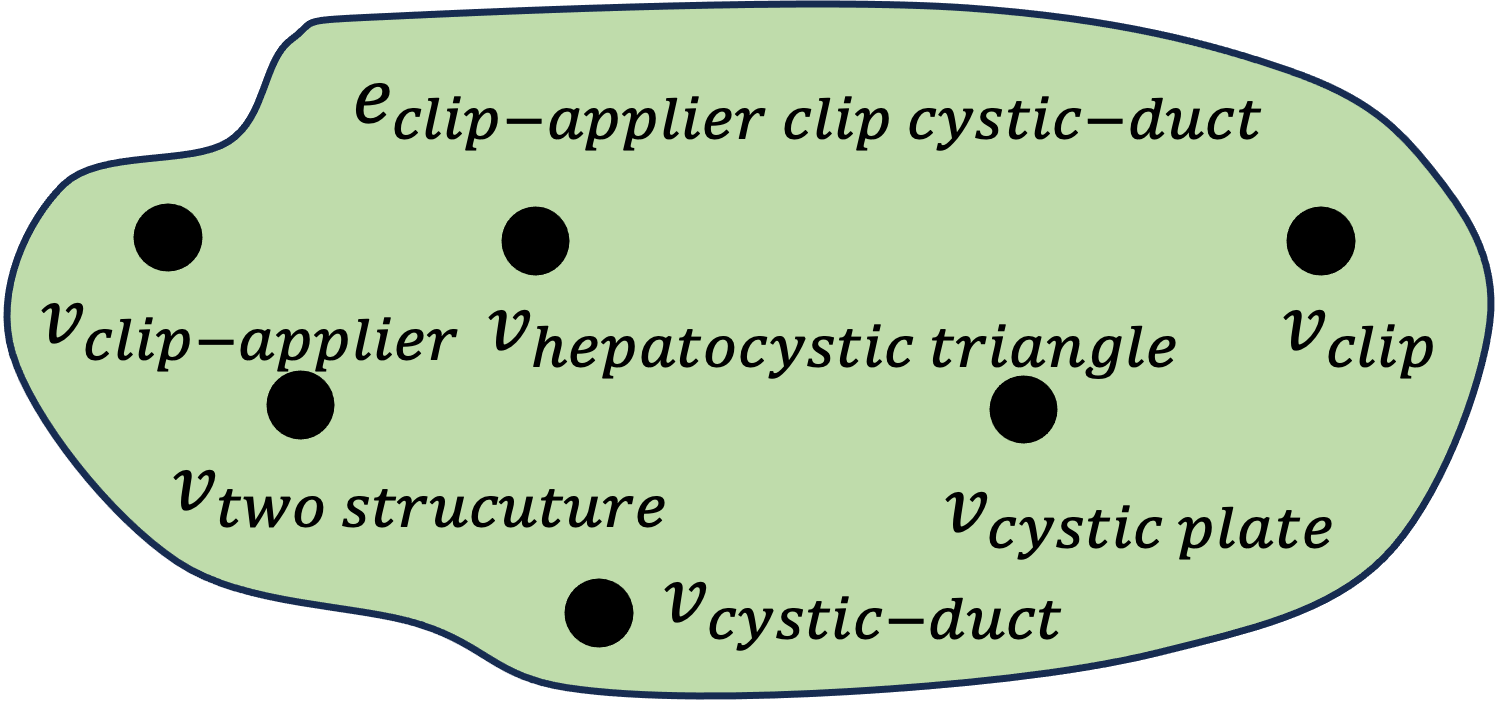}

    (d)
    \label{fig:node_edge for clipping}
    \end{minipage}  

    \caption{Subgraphs for different prediction tasks: a) action-triplets. b) CVS achievement. c) Clipping without prior CVS. d) Clipping.}
    \label{fig:subgraphs}
\end{figure*}
\section{Experimental setup} \label{se:experimental_setup}
\subsection{Clinical Problems}
To verify the effectiveness of the proposed framework, we set up three experiments, which are predicting action-triplet, CVS progression, and clipping of the cystic duct and cystic artery, especially in the absence of CVS verification. 

As predicting the tools, tissues, action, and action-triplet during laparoscopic cholecystectomy indicates a detailed prediction on the imminent steps in surgery as explained in the introduction. With accurate detection and prediction, one can give warnings before a wrong action happens. The tools in the videos for the paper are taken from \cite{Nwoye2020-qm}, and include $\textit{clip-applier}$,$\textit{irrigator}$, $\textit{scissors}$, $\textit{bipolar}$, $\textit{hook}$. 
Target tissues include $\textit{liver}$, $\textit{cystic-duct}$, $\textit{cystic-artery}$, $\textit{gallbladder}$, and others. Actions concepts include $\textit{clip}$, $\textit{grasp}$, $\textit{cut}$ and so on. There are 100 action-triplet pairs to be predicted overall. 

Critical view of safety as explained in introduction is a required safety check before proceeding to clipping of cystic duct and cystic artery. Prediction on whether such criteria are achieved during surgery provides a valuable indication to the surgeon on the safety of the surgery. 

Action-triplet involving clip-applier clip cystic duct and cystic artery is irreversible actions. The common mistake is to clip or injure the common bile duct. One of the safety checks to avoid such mistakes is to check if CVS is achieved before the action. The goal is to predict these two action-triplets of clip-applier clip cystic duct and cystic artery. 


\subsection{Datasets}
For the action-triplets dataset in the first and third experiments, we used the publicly available CholecT45\cite{Nwoye2022-cr} dataset which is a subset of Cholec80 \cite{Twinanda2017-sj}. It consists of 45 videos including annotations for 6 surgical tools, 15 tissues, 10 actions, and 100 action-triplets. We used a sampling rate of 1 frame per second (FPS) for the input video. For the prediction problem we used frame from past 4 second. For the detection ablation study, we used the current frame and no past frames. We used 80\% of the data for training and 20\% for validation. 
For the CVS dataset in the second and third experiments, we used the Cholec80-CVS dataset \cite{Rios2023-hz} and follow the labeling from there, with concepts of 
$\textit{two-structures}$,
$\textit{cystic-plate}$,
$\textit{hepatocystic-triangle}$, and $\textit{CVS-achieved}$ as a 3-way relation connecting them.

\subsection{Comparison Methods} 
The comparison of the methods are Tripnet \cite{Nwoye2020-ou}, attention Tripnet~\cite{Nwoye2022-cr}, Rendezvous method~\cite{Nwoye2022-cr}, and an encoder-decoder \cite{cho2014properties} recurrent state space models (RSSM) ~\cite{Hafner2020-zz}, hypergraph with LSTM (Graph LSTM). The attention tripnet, tripnet, and Rendezvous were explained in related work. The recurrent state space models enforce the prediction of the hidden state for the future states and, in return, train an encoder that can represent the future hidden state better. Hypergraph with LSTM (Graph LSTM) uses the same hypergraph structure as the proposed method but uses LSTM instead of a transformer for prediction of node and edge embedding. 

\section{Results} \label{se: results}

\subsection{CholecT45}
We begin by demonstrating the performance of the model to predict action-triplets in the future. For each action-triplet, the triplet is represented by the hypergraph edge; The associated tool, action, and tissue are represented by graph nodes (Fig.~\ref{fig:subgraphs}(a)). Since the true positive rate of action-triplet is low during evaluation, the average precision (AP) \cite{Nwoye2022-cr} is used as one of the metrics to evaluate the performance of the different methods. 

In the experiment, we keep the prediction horizon (shortened as hp in the table) at 4 seconds, as it is a long enough horizon for the surgeon to intervene if some high-risk events are predicted. Prediction is a harder task compared to recognition of the current state.  However, the proposed model does not suffer from a significant drop, reduced by 5.59\% from 30.75\% to 25.16\% while predicting 4 seconds into future with 4 second past frames (Tab.~\ref{tb:triplet_prediction}). We also show the proposed HGT achieves the best prediction, thanks to the causal transformer design.
 
We also did an ablation study to demonstrate the performance of action-triplet detection at the current time point. The detection performance is also related to backbone selection and pretraining which is not the paper's main focus. The comparison focuses on the related work using the backbone of ResNet and ViT with similar number of parameters, since performances of them are comparable in this task shown in \cite{Ban2023-ey}. The AP of action-triplet recognition is 30.05\% for the proposed HGT with only current frame, and is comparable to the state-of-the-art performance compared to baselines such as Tripnet \cite{Nwoye2020-ou} (24.4\%), Rendezvous \cite{Nwoye2022-cr} (29.4\%), and Attention tripnet \cite{Nwoye2022-cr} (27.2\%) with only current frame as input. The AP of action-triplet recognition is 30.75 \% for the proposed HGT with both past and current frame, which is also comparable to the results of RiT \cite{sharma2023rendezvous} (29.7\%) (Tab.~\ref{tb:triplet_prediction}).



\begin{table}[ht]
    \centering
    \caption{Action-triplets. }
    \begin{tabular}{ccc}
    \hline 
            & \multicolumn{2}{c}{AP of action-triplet }   \\
    \hline  Method                                  & hp=0                 & hp=4s                     \\
    \hline  Tripnet \cite{Nwoye2020-ou}             & 24.4                 & -                         \\
    \hline  Rendezvous \cite{Nwoye2022-cr}          & 29.4                 & -                         \\
    \hline  Attention tripnet \cite{Nwoye2022-cr}   &27.2                  & -                         \\
    \hline  RiT \cite{sharma2023rendezvous}         &29.7                  & - 
    \\
    \hline  RSSM~\cite{Hafner2020-zz}               & 26.46                & 20.40                     \\
    \hline  Graph LSTM                              & 28.46                &  22.25                    \\
    \hline  \textbf{HGT (proposed)}                 & \textbf{30.75}       &\textbf{25.16}             \\
    \hline
    \end{tabular}
    \label{tb:triplet_prediction} 
\end{table}

\subsection{CVS Progression}
According to the medical definition of CVS problem (shown in Fig~\ref{fig:subgraphs}(b)), the nodes represent the three criteria of the CVS and the edge represents the relations between these criteria.
The main motivation for using the Cholec80-CVS dataset is to be consistent with the other two experiments in the paper and public accessibility. 
We show results for CVS progression estimates in Tab~\ref{tb:cvs_detection and prediction }. We keep the prediction horizon at 4 seconds to be consistent. The accuracy is reduced by only 3.1\% in the 4-second prediction horizon compared to detection at the current time point. 
We also show that the proposed method in CVS detection is comparable to state-of-the-art performance at 0-second detection task (Tab~\ref{tb:cvs_detection and prediction }) as an ablation study.

\begin{table}[ht]
    \centering
    \caption{CVS detection and prediction average accuracy}
    \begin{tabular}{ccc}
    \hline
             \multicolumn{3}{c}{CVS Accuracy \%}  \\
    \hline 
            Method                      & hp=0          & hp=4s   \\
    \hline
            MTL \cite{Nwoye2020-ou}     & 52.3          & -       \\
    \hline
            Tripnet \cite{Nwoye2020-ou} & 50.1          & -        \\
    \hline
            I3D \cite{Carreira2017-tv}  & 49.9          & -       \\ 
    \hline 
            CSN \cite{Pan2019-hh}       & 50.0          & -       \\ 
    \hline 
            TSM \cite{Lin2020-jl}       & 57.7          & -       \\
    \hline  Med-GEMINI \cite{saab2024capabilities} &55.2 & -    \\
    \hline
            RSSM~\cite{Hafner2020-zz}                        & 50.87         & 49.3  \\
    \hline
            Graph LSTM                  & 52.67         & 48.07   \\
    \hline 
            \textbf{HGT (proposed)}                     &\textbf{60.66} &\textbf{57.56}   \\
    \hline
    \end{tabular}
    \label{tb:cvs_detection and prediction } 
\end{table}

\subsection{Clipping Prediction}
In the clipping prediction task, the edge represents the clip-applier clip cystic duct and cystic artery, and the nodes represent clip-applier, clip, cystic duct, and cystic artery, as shown in Fig.~\ref{fig:subgraphs}(c). To include the prior knowledge from human experts, we include the overall CVS as well as the three individual CVS criteria into the nodes, as shown in Fig.~\ref{fig:subgraphs}(d). We find that with prior knowledge the average precision of clip-applier clip cystic-duct and cystic-artery improved from 51.80\% to 56.02\% in the prediction horizon of 4 seconds and from 62.68\% to 67.41\% in detection (prediction horizon of 0 seconds) in Tab~\ref{tb: clip prediction}. The ablation study also shows that our approach reaches state-of-the-art in detection (Tab.~\ref{tb: clip prediction})\footnote{Note that $<$ 46 in the table indicates that 46\% is the 10th highest of AP of all 100 action-triplets shown in the paper, and the clip-applier clip cystic-duct and cystic-artery are below this value~\cite{Nwoye2022-cr}}.


\begin{table}[ht]
    \centering
    \caption{clip-applier clip cystic-duct and clip-applier clip cystic-artery}
    \begin{tabular}{ccccc}
            \hline 
            & \multicolumn{4}{c}{Average precision}                                 \\\hline
            Method                               & hp=0  & hp=4s \\\hline 
            Tripnet \cite{Nwoye2020-ou}          & 43.53 & -     \\\hline
            Rendezvous \cite{Nwoye2022-cr}       & 55.6  & -     \\\hline
            Attention Tripnet \cite{Nwoye2022-cr}& $<$ 46& -     \\\hline 
            RSSM~\cite{Hafner2020-zz}            & 60.69 & 50.33 \\\hline 
            Graph LSTM                           & 57.58 & 47.93 \\\hline 
            HGT 
            $w \backslash o$ 
            CVS prior
                                                 & 62.68 & 51.80 \\\hline 
            \textbf{HGT (proposed)}              & \textbf{67.41}  & \textbf{56.02}  \\\hline 
    \end{tabular}
    \label{tb: clip prediction} 
\end{table}

The action of achieving CVS and clipping is positively correlated (Eq.~\ref{eq: correlation}), where $e$ represents the edge of $\langle\textit{clip-applier,clip,cystic-duct}\rangle$ and $v$ represents the nodes of $\textit{clip-applier}$, $\textit{clip}$, $\textit{cystic-duct}$, $\textit{CVS}$, and $\textit{two-structures}$, $\textit{cystic-plate}$, $\textit{hepatocystic-triangle}$. 
This leads to the fact that the conditioned probability $P(e|v) $ is higher than $P(e)$. 
With the introduction of prior evidence in nodes, the accuracy of detection of interest action-triplets and concepts increased in maximum-likelihood estimation during training. 
\begin{equation} \label{eq: correlation}
\begin{aligned}
 \rho & \propto (P(e,v) - P(e)P(v)) \\
& = P(v)(P(e|v) - P(e)) \\
& \propto P(e|v) - P(e)
\end{aligned}
\end{equation}

Clipping in the absence of CVS achievement can lead to significant complications, such as common bile duct injuries. Therefore, we evaluate the model's performance in this case. We find that the average precision of clipping-triplet detection is 72.32\% and the average precision of clipping-triplet prediction for 4-second is 59.80\% in the absence of CVS achievement. This gives a strong indication that the proposed hypergraph-transformer (HGT) can provide clinically relevant information to the surgeon when safety criteria (CVS) are not achieved in laparoscopic cholecystectomy. 

\section{Discussion and limitation} \label{se: discussion and limitation}
As with other domains where prediction is applied to safety systems, prediction is limited by the low abundance of targeted adverse events in the training data. Similarly, our datasets do not contain adverse events relating to surgical complications in a sufficient quantity to guarantee adequate ML performance on out-of-distribution samples. Predicting major surgical complications or errors, such as the clip-applier clipping the common bile duct instead of the cystic duct or cystic artery, is a challenging endeavour as such events are not frequent in real-world datasets. This phenomenon has motivated extensive research in confidence estimates for prediction approaches \cite{Huang2019-ic,Huang2021-fb,huang2022tip,Luo2021-rj,Ettinger2021-zr,Nayak2022-xw, Richter2017-ca,Fridovich-Keil2020-qw,Fisac2018-yx}, and remains an unsolved problem. On the contrary, prediction of correct action can ensure the prevention of adverse events or complications, thus indirectly enhancing the safety provided by assistive interfaces. Overall, there are more capabilities to explore to achieve a holistic safety and warning system for complex and risk-prone settings, such as MIRS, which is beyond the scope of this paper.

The small size of the dataset limits inference accuracy for complex, multi-variable models to describe the achievement of CVS and the individual criteria, even for balanced datasets. The relative abundance of CVS achievement and of the individual CVS criteria further exacerbates the problem when exploring this important surgical safety measure. 
Moreover, annotation of the CVS is inherently subjective and not standardized \cite{Ward2021-va}, which limits the interpretability and affects the use of the prediction in downstream applications towards clinical use. We focus on the prediction capability of the proposed structure and do not investigate the problem of dataset abundance and annotation variability further. 

Another factor influencing the clinical relevance of such approaches is the relation between the offered prediction horizon and the downstream interaction between the assistive system and the surgeon. The ideal predictive horizon has not been determined yet, and is known to be crucial for warning systems in other fields~\cite{petermeijer2017driver,mehrotra2022human}. Considering that human reaction time to process an image is under 150 ms \cite{thorpe1996speed}, our prediction horizon is an order of magnitude longer than that. While our prediction horizons were chosen in consultation with subject matter experts, further human-in-the-loop investigations specifically in the surgical setting are needed to establish validated criteria for these horizons.

\section{Conclusion} \label{se: conclusion}
We show the performance of the proposed hypergraph-based prediction method in the prediction of three clinically relevant tasks. Compared to reference methods, the proposed method has superior performance in the prediction of critical events, such as clipping of cystic duct or artery, without prior CVS achievement in the 4-second prediction horizon. The average precision decreased by only 5.59\% between action-triplet detection and prediction on time horizons relevant to an assistive system.
In conclusion, the proposed framework can give adequate predictions of surgical events and actions, thus providing a relevant component of future robotic and AI-enabled assistive systems in surgery. 

\bibliographystyle{IEEEtran}
\bibliography{paperpile}  

\begin{thebibliography}{10}
\providecommand{\url}[1]{#1}
\csname url@samestyle\endcsname
\providecommand{\newblock}{\relax}
\providecommand{\bibinfo}[2]{#2}
\providecommand{\BIBentrySTDinterwordspacing}{\spaceskip=0pt\relax}
\providecommand{\BIBentryALTinterwordstretchfactor}{4}
\providecommand{\BIBentryALTinterwordspacing}{\spaceskip=\fontdimen2\font plus
\BIBentryALTinterwordstretchfactor\fontdimen3\font minus \fontdimen4\font\relax}
\providecommand{\BIBforeignlanguage}[2]{{%
\expandafter\ifx\csname l@#1\endcsname\relax
\typeout{** WARNING: IEEEtran.bst: No hyphenation pattern has been}%
\typeout{** loaded for the language `#1'. Using the pattern for}%
\typeout{** the default language instead.}%
\else
\language=\csname l@#1\endcsname
\fi
#2}}
\providecommand{\BIBdecl}{\relax}
\BIBdecl

\bibitem{Ivanovic2020-ns}
B.~Ivanovic, A.~Elhafsi, G.~Rosman, A.~Gaidon, and {others}, ``Mats: An interpretable trajectory forecasting representation for planning and control,'' \emph{arXiv preprint arXiv}, 2020.

\bibitem{Balachandran2023-lf}
A.~Balachandran, T.~L. Chen, J.~Y.~M. Goh, S.~McGill, G.~Rosman, S.~Stent, and J.~J. Leonard, ``Human-centric intelligent driving: Collaborating with the driver to improve safety,'' in \emph{Road Vehicle Automation 9}.\hskip 1em plus 0.5em minus 0.4em\relax Springer International Publishing, 2023, pp. 85--109.

\bibitem{Rudenko2020-cp}
A.~Rudenko, L.~Palmieri, M.~Herman, K.~M. Kitani, D.~M. Gavrila, and K.~O. Arras, ``Human motion trajectory prediction: a survey,'' \emph{Int. J. Rob. Res.}, vol.~39, no.~8, pp. 895--935, Jul. 2020.

\bibitem{Truong2018-mp}
X.-T. Truong and T.-D. Ngo, ````to approach humans?'': A unified framework for approaching pose prediction and socially aware robot navigation,'' \emph{IEEE Transactions on Cognitive and Developmental Systems}, vol.~10, no.~3, pp. 557--572, Sep. 2018.

\bibitem{Mavrogiannis2023-cp}
C.~Mavrogiannis, F.~Baldini, A.~Wang, D.~Zhao, P.~Trautman, A.~Steinfeld, and J.~Oh, ``Core challenges of social robot navigation: A survey,'' \emph{J. Hum.-Robot Interact.}, vol.~12, no.~3, pp. 1--39, Apr. 2023.

\bibitem{Fridovich-Keil2020-qw}
D.~Fridovich-Keil, A.~Bajcsy, J.~F. Fisac, S.~L. Herbert, S.~Wang, A.~D. Dragan, and C.~J. Tomlin, ``Confidence-aware motion prediction for real-time collision avoidance1,'' \emph{Int. J. Rob. Res.}, vol.~39, no. 2-3, pp. 250--265, Mar. 2020.

\bibitem{Kanazawa_undated-cj}
A.~Kanazawa, {Member}, J.~Kinugawa, {Member}, and K.~Kosuge, ``Adaptive motion planning for a collaborative robot based on prediction uncertainty to enhance human safety and work efficiency,'' \emph{T-RO}, 2019.

\bibitem{cai2023prediction}
J.~Cai, A.~Du, X.~Liang, and S.~Li, ``Prediction-based path planning for safe and efficient human--robot collaboration in construction via deep reinforcement learning,'' \emph{Journal of computing in civil engineering}, vol.~37, no.~1, p. 04022046, 2023.

\bibitem{li2024safe}
W.~Li, Y.~Hu, Y.~Zhou, and D.~T. Pham, ``Safe human--robot collaboration for industrial settings: a survey,'' \emph{Journal of Intelligent Manufacturing}, vol.~35, no.~5, pp. 2235--2261, 2024.

\bibitem{cameron1991laparoscopic}
J.~Cameron and T.~Gadacz, ``Laparoscopic cholecystectomy.'' \emph{Annals of Surgery}, vol. 213, no.~1, p.~1, 1991.

\bibitem{Soper1991-bb}
N.~J. Soper, ``\BIBforeignlanguage{en}{Laparoscopic cholecystectomy},'' \emph{\BIBforeignlanguage{en}{Curr. Probl. Surg.}}, vol.~28, no.~9, pp. 581--655, Sep. 1991.

\bibitem{Vollmer2007-ri}
C.~M. Vollmer, Jr and M.~P. Callery, ``\BIBforeignlanguage{en}{Biliary injury following laparoscopic cholecystectomy: why still a problem?}'' \emph{\BIBforeignlanguage{en}{Gastroenterology}}, vol. 133, no.~3, pp. 1039--1041, Sep. 2007.

\bibitem{Nwoye2020-qm}
C.~I. Nwoye, C.~Gonzalez, T.~Yu, P.~Mascagni, D.~Mutter, J.~Marescaux, and N.~Padoy, ``Recognition of {Instrument-Tissue} interactions in endoscopic videos via action triplets,'' in \emph{Medical Image Computing and Computer Assisted Intervention -- {MICCAI} 2020}.\hskip 1em plus 0.5em minus 0.4em\relax Springer International Publishing, 2020, pp. 364--374.

\bibitem{Nwoye2023-ci}
C.~I. Nwoye, D.~Alapatt, T.~Yu, A.~Vardazaryan, F.~Xia, Z.~Zhao, T.~Xia, F.~Jia, Y.~Yang, H.~Wang, D.~Yu, G.~Zheng, X.~Duan, N.~Getty, R.~Sanchez-Matilla, M.~Robu, L.~Zhang, H.~Chen, J.~Wang, L.~Wang, B.~Zhang, B.~Gerats, S.~Raviteja, R.~Sathish, R.~Tao, S.~Kondo, W.~Pang, H.~Ren, J.~R. Abbing, M.~H. Sarhan, S.~Bodenstedt, N.~Bhasker, B.~Oliveira, H.~R. Torres, L.~Ling, F.~Gaida, T.~Czempiel, J.~L. Vila{\c c}a, P.~Morais, J.~Fonseca, R.~M. Egging, I.~N. Wijma, C.~Qian, G.~Bian, Z.~Li, V.~Balasubramanian, D.~Sheet, I.~Luengo, Y.~Zhu, S.~Ding, J.-A. Aschenbrenner, N.~E. van~der Kar, M.~Xu, M.~Islam, L.~Seenivasan, A.~Jenke, D.~Stoyanov, D.~Mutter, P.~Mascagni, B.~Seeliger, C.~Gonzalez, and N.~Padoy, ``\BIBforeignlanguage{en}{{CholecTriplet2021}: A benchmark challenge for surgical action triplet recognition},'' \emph{\BIBforeignlanguage{en}{Med. Image Anal.}}, vol.~86, p. 102803, May 2023.

\bibitem{Strasberg1995-jy}
S.~M. Strasberg, M.~Hertl, and N.~J. Soper, ``\BIBforeignlanguage{en}{An analysis of the problem of biliary injury during laparoscopic cholecystectomy},'' \emph{\BIBforeignlanguage{en}{J. Am. Coll. Surg.}}, vol. 180, no.~1, pp. 101--125, Jan. 1995.

\bibitem{Strasberg2010-eq}
S.~M. Strasberg and L.~M. Brunt, ``\BIBforeignlanguage{en}{Rationale and use of the critical view of safety in laparoscopic cholecystectomy},'' \emph{\BIBforeignlanguage{en}{J. Am. Coll. Surg.}}, vol. 211, no.~1, pp. 132--138, Jul. 2010.

\bibitem{Ulutan2020-mf}
O.~Ulutan, A.~S.~M. Iftekhar, and B.~S. Manjunath, ``{VSGNet}: Spatial attention network for detecting human object interactions using graph convolutions,'' \emph{arXiv}, pp. 13\,617--13\,626, Mar. 2020.

\bibitem{Maraghi2021-tb}
V.~O. Maraghi and K.~Faez, ``\BIBforeignlanguage{en}{Scaling {Human-Object} interaction recognition in the video through {Zero-Shot} learning},'' \emph{\BIBforeignlanguage{en}{Comput. Intell. Neurosci.}}, vol. 2021, p. 9922697, Jun. 2021.

\bibitem{Rupprecht2016-si}
C.~Rupprecht, C.~Lea, F.~Tombari, N.~Navab, and G.~D. Hager, ``Sensor substitution for video-based action recognition,'' in \emph{IROS}.\hskip 1em plus 0.5em minus 0.4em\relax IEEE, Oct. 2016, pp. 5230--5237.

\bibitem{Mallya2016-ip}
A.~Mallya and S.~Lazebnik, ``Learning models for actions and {Person-Object} interactions with transfer to question answering,'' in \emph{Computer Vision -- {ECCV} 2016}.\hskip 1em plus 0.5em minus 0.4em\relax Springer International Publishing, 2016, pp. 414--428.

\bibitem{Ban2023-ey}
Y.~Ban, J.~A. Eckhoff, T.~M. Ward, D.~A. Hashimoto, O.~R. Meireles, D.~Rus, and G.~Rosman, ``\BIBforeignlanguage{en}{Concept graph neural networks for surgical video understanding},'' \emph{\BIBforeignlanguage{en}{IEEE Trans. Med. Imaging}}, Jul. 2023.

\bibitem{Hochreiter1997-gm}
S.~Hochreiter and J.~Schmidhuber, ``\BIBforeignlanguage{en}{Long short-term memory},'' \emph{\BIBforeignlanguage{en}{Neural Comput.}}, vol.~9, no.~8, pp. 1735--1780, Nov. 1997.

\bibitem{Salzmann2020-pg}
T.~Salzmann, B.~Ivanovic, P.~Chakravarty, and M.~Pavone, ``Trajectron++: {Dynamically-Feasible} trajectory forecasting with heterogeneous data,'' in \emph{Computer Vision -- {ECCV} 2020}.\hskip 1em plus 0.5em minus 0.4em\relax Springer International Publishing, 2020, pp. 683--700.

\bibitem{Alahi2016-zq}
A.~Alahi, K.~Goel, V.~Ramanathan, A.~Robicquet, L.~Fei-Fei, and S.~Savarese, ``Social {LSTM}: Human trajectory prediction in crowded spaces,'' in \emph{CVPR}, 2016, pp. 961--971.

\bibitem{Mohamed2020-hu}
A.~Mohamed, K.~Qian, M.~Elhoseiny, and {others}, ``Social-{STGCNN}: A social spatio-temporal graph convolutional neural network for human trajectory prediction,'' in \emph{CVPR}.\hskip 1em plus 0.5em minus 0.4em\relax openaccess.thecvf.com, 2020.

\bibitem{Li2020-jp}
J.~Li, F.~Yang, M.~Tomizuka, and C.~Choi, ``{EvolveGraph}: Multi-agent trajectory prediction with dynamic relational reasoning,'' in \emph{NeurIPS}, Mar. 2020.

\bibitem{Vaswani2017-bn}
A.~Vaswani, N.~Shazeer, N.~Parmar, J.~Uszkoreit, L.~Jones, A.~N. Gomez, {\L}.~Kaiser, and I.~Polosukhin, ``Attention is all you need,'' \emph{Adv. Neural Inf. Process. Syst.}, vol.~30, 2017.

\bibitem{Ngiam2022-ny}
J.~Ngiam, B.~Caine, V.~Vasudevan, Z.~Zhang, H.-T.~L. Chiang, J.~Ling, R.~Roelofs, A.~Bewley, C.~Liu, A.~Venugopal, D.~Weiss, B.~Sapp, Z.~Chen, and J.~Shlens, ``Scene transformer: A unified architecture for predicting multiple agent trajectories,'' in \emph{ICLR}, 2022.

\bibitem{sharma2023rendezvous}
S.~Sharma, C.~I. Nwoye, D.~Mutter, and N.~Padoy, ``Rendezvous in time: an attention-based temporal fusion approach for surgical triplet recognition,'' \emph{International Journal of Computer Assisted Radiology and Surgery}, vol.~18, no.~6, pp. 1053--1059, 2023.

\bibitem{Goodfellow_undated-qk}
I.~J. Goodfellow, J.~Pouget-Abadie, M.~Mirza, B.~Xu, D.~Warde-Farley, S.~Ozair, A.~Courville, and Y.~Bengio, ``Generative adversarial nets,'' in \emph{NIPS}, 2014, accessed: 2024-1-12.

\bibitem{Creswell2018-ca}
A.~Creswell, T.~White, V.~Dumoulin, K.~Arulkumaran, B.~Sengupta, and A.~A. Bharath, ``Generative adversarial networks: An overview,'' \emph{IEEE Signal Process. Mag.}, vol.~35, no.~1, pp. 53--65, Jan. 2018.

\bibitem{Liang2017-sd}
X.~Liang, L.~Lee, W.~Dai, and E.~P. Xing, ``Dual motion {GAN} for future-flow embedded video prediction,'' in \emph{ICCV}.\hskip 1em plus 0.5em minus 0.4em\relax IEEE, Oct. 2017.

\bibitem{Ban2022-gl}
Y.~Ban, G.~Rosman, J.~A. Eckhoff, T.~M. Ward, D.~A. Hashimoto, T.~Kondo, H.~Iwaki, O.~R. Meireles, and D.~Rus, ``{SUPR-GAN}: {SUrgical} {PRediction} {GAN} for event anticipation in laparoscopic and robotic surgery,'' \emph{RA-L}, vol.~7, no.~2, pp. 5741--5748, Apr. 2022.

\bibitem{Gupta2018-go}
A.~Gupta, J.~Johnson, L.~Fei-Fei, S.~Savarese, and A.~Alahi, ``Social {GAN}: Socially acceptable trajectories with generative adversarial networks,'' in \emph{CVPR}, Mar. 2018.

\bibitem{Huang2020-pr}
X.~Huang, S.~G. McGill, J.~A. DeCastro, L.~Fletcher, J.~J. Leonard, B.~C. Williams, and G.~Rosman, ``{DiversityGAN}: {Diversity-Aware} vehicle motion prediction via latent semantic sampling,'' \emph{RA-L}, vol.~5, no.~4, pp. 5089--5096, Oct. 2020.

\bibitem{Scarselli2009-tl}
F.~Scarselli, M.~Gori, A.~C. Tsoi, M.~Hagenbuchner, and G.~Monfardini, ``\BIBforeignlanguage{en}{The graph neural network model},'' \emph{\BIBforeignlanguage{en}{IEEE Trans. Neural Netw.}}, vol.~20, no.~1, pp. 61--80, Jan. 2009.

\bibitem{Murali2024-at}
A.~Murali, D.~Alapatt, P.~Mascagni, A.~Vardazaryan, A.~Garcia, N.~Okamoto, D.~Mutter, and N.~Padoy, ``\BIBforeignlanguage{en}{Latent graph representations for critical view of safety assessment},'' \emph{\BIBforeignlanguage{en}{IEEE Trans. Med. Imaging}}, vol.~43, no.~3, pp. 1247--1258, Mar. 2024.

\bibitem{Xu2020-of}
K.~Xu, M.~Zhang, J.~Li, S.~S. Du, K.-I. Kawarabayashi, and S.~Jegelka, ``How neural networks extrapolate: From feedforward to graph neural networks,'' in \emph{ICLR}, Sep. 2020.

\bibitem{Feng2019-wz}
Y.~Feng, H.~You, Z.~Zhang, R.~Ji, and Y.~Gao, ``\BIBforeignlanguage{en}{Hypergraph neural networks},'' \emph{\BIBforeignlanguage{en}{AAAI}}, vol.~33, no.~01, pp. 3558--3565, Jul. 2019.

\bibitem{bretto:hal-01024351}
\BIBentryALTinterwordspacing
A.~Bretto, \emph{{Hypergraph Theory - An Introduction}}.\hskip 1em plus 0.5em minus 0.4em\relax {Springer International Publishing}, 2013. [Online]. Available: \url{https://hal.archives-ouvertes.fr/hal-01024351}
\BIBentrySTDinterwordspacing

\bibitem{Roy2020-is}
M.~Roy, L.~Montorfano, and R.~J. Rosenthal, ``Laparoscopic cholecystectomy,'' in \emph{Mental Conditioning to Perform Common Operations in General Surgery Training: A Systematic Approach to Expediting Skill Acquisition and Maintaining Dexterity in Performance}, R.~J. Rosenthal, A.~Rosales, E.~Lo~Menzo, and F.~D. Dip, Eds.\hskip 1em plus 0.5em minus 0.4em\relax Cham: Springer International Publishing, 2020, pp. 153--158.

\bibitem{Dosovitskiy2020-dp}
A.~Dosovitskiy, L.~Beyer, A.~Kolesnikov, D.~Weissenborn, X.~Zhai, T.~Unterthiner, M.~Dehghani, M.~Minderer, G.~Heigold, S.~Gelly, J.~Uszkoreit, and N.~Houlsby, ``An image is worth 16x16 words: Transformers for image recognition at scale,'' in \emph{ICLR}, Oct. 2021.

\bibitem{Caron2021-rb}
M.~Caron, H.~Touvron, I.~Misra, H.~Jegou, J.~Mairal, P.~Bojanowski, and A.~Joulin, ``Emerging properties in self-supervised vision transformers,'' in \emph{ICCV}.\hskip 1em plus 0.5em minus 0.4em\relax IEEE, Oct. 2021, pp. 9650--9660.

\bibitem{Velickovic2017-xf}
P.~Veli{\v c}kovi{\'c}, G.~Cucurull, A.~Casanova, A.~Romero, P.~Li{\`o}, and Y.~Bengio, ``Graph attention networks,'' \emph{arXiv}, Oct. 2017.

\bibitem{Ngiam2021-hn}
J.~Ngiam, B.~Caine, V.~Vasudevan, Z.~Zhang, H.-T.~L. Chiang, J.~Ling, R.~Roelofs, A.~Bewley, C.~Liu, A.~Venugopal, D.~Weiss, B.~Sapp, Z.~Chen, and J.~Shlens, ``Scene transformer: A unified architecture for predicting multiple agent trajectories,'' in \emph{ICLR}, Jun. 2022.

\bibitem{Lin2020-of}
T.-Y. Lin, P.~Goyal, R.~Girshick, K.~He, and P.~Dollar, ``\BIBforeignlanguage{en}{Focal loss for dense object detection},'' \emph{\BIBforeignlanguage{en}{IEEE Trans. Pattern Anal. Mach. Intell.}}, vol.~42, no.~2, pp. 318--327, Feb. 2020.

\bibitem{Nwoye2022-cr}
C.~I. Nwoye, T.~Yu, C.~Gonzalez, B.~Seeliger, P.~Mascagni, D.~Mutter, J.~Marescaux, and N.~Padoy, ``\BIBforeignlanguage{en}{Rendezvous: Attention mechanisms for the recognition of surgical action triplets in endoscopic videos},'' \emph{\BIBforeignlanguage{en}{Med. Image Anal.}}, vol.~78, p. 102433, May 2022.

\bibitem{Twinanda2017-sj}
A.~P. Twinanda, S.~Shehata, D.~Mutter, J.~Marescaux, M.~de~Mathelin, and N.~Padoy, ``\BIBforeignlanguage{en}{{EndoNet}: A deep architecture for recognition tasks on laparoscopic videos},'' \emph{\BIBforeignlanguage{en}{IEEE Trans. Med. Imaging}}, vol.~36, no.~1, pp. 86--97, Jan. 2017.

\bibitem{Rios2023-hz}
M.~S. R{\'\i}os, M.~A. Molina-Rodriguez, D.~Londo{\~n}o, C.~A. Guill{\'e}n, S.~Sierra, F.~Zapata, and L.~F. Giraldo, ``\BIBforeignlanguage{en}{{Cholec80-CVS}: An open dataset with an evaluation of strasberg's critical view of safety for {AI}},'' \emph{\BIBforeignlanguage{en}{Sci Data}}, vol.~10, no.~1, p. 194, Apr. 2023.

\bibitem{Nwoye2020-ou}
C.~I. Nwoye, C.~Gonzalez, T.~Yu, P.~Mascagni, D.~Mutter, J.~Marescaux, and N.~Padoy, ``Recognition of {Instrument-Tissue} interactions in endoscopic videos via action triplets,'' in \emph{MICCAI}, Jul. 2020.

\bibitem{cho2014properties}
K.~Cho, B.~van Merri{\"e}nboer, D.~Bahdanau, and Y.~Bengio, ``On the properties of neural machine translation: Encoder--decoder approaches,'' \emph{Syntax, Semantics and Structure in Statistical Translation}, p. 103, 2014.

\bibitem{Hafner2020-zz}
D.~Hafner, T.~Lillicrap, M.~Norouzi, and J.~Ba, ``Mastering atari with discrete world models,'' in \emph{ICLR}, Oct. 2020.

\bibitem{Carreira2017-tv}
J.~Carreira and A.~Zisserman, ``Quo vadis, action recognition? a new model and the kinetics dataset,'' in \emph{CVPR}, May 2017, pp. 6299--6308.

\bibitem{Pan2019-hh}
B.~Pan, J.~Sun, W.~Lin, L.~Wang, and W.~Lin, ``Cross-stream selective networks for action recognition,'' in \emph{CVPR Workshops}.\hskip 1em plus 0.5em minus 0.4em\relax IEEE, Jun. 2019, pp. 0--0.

\bibitem{Lin2020-jl}
J.~Lin, C.~Gan, K.~Wang, and S.~Han, ``{TSM}: Temporal shift module for efficient and scalable video understanding on edge devices,'' \emph{IEEE Trans. Pattern Anal. Mach. Intell.}, pp. 1--1, 2020.

\bibitem{saab2024capabilities}
K.~Saab, T.~Tu, W.-H. Weng, R.~Tanno, D.~Stutz, E.~Wulczyn, F.~Zhang, T.~Strother, C.~Park, E.~Vedadi \emph{et~al.}, ``Capabilities of gemini models in medicine,'' \emph{arXiv preprint arXiv:2404.18416}, 2024.

\bibitem{Huang2019-ic}
X.~Huang, S.~G. McGill, B.~C. Williams, L.~Fletcher, and G.~Rosman, ``{Uncertainty-Aware} driver trajectory prediction at urban intersections,'' in \emph{ICRA}, May 2019, pp. 9718--9724.

\bibitem{Huang2021-fb}
X.~Huang, S.~G. McGill, J.~A. DeCastro, L.~Fletcher, J.~J. Leonard, B.~C. Williams, and G.~Rosman, ``{CARPAL}: {Confidence-Aware} intent recognition for parallel autonomy,'' \emph{RA-L}, 2021.

\bibitem{huang2022tip}
X.~Huang, G.~Rosman, A.~Jasour, S.~G. McGill, J.~J. Leonard, and B.~C. Williams, ``Tip: Task-informed motion prediction for intelligent vehicles,'' in \emph{IROS}.\hskip 1em plus 0.5em minus 0.4em\relax IEEE, 2022, pp. 11\,432--11\,439.

\bibitem{Luo2021-rj}
R.~Luo, S.~Zhao, J.~Kuck, B.~Ivanovic, S.~Savarese, E.~Schmerling, and M.~Pavone, ``{Sample-Efficient} safety assurances using conformal prediction,'' \emph{arXiv}, Sep. 2021.

\bibitem{Ettinger2021-zr}
S.~Ettinger, S.~Cheng, B.~Caine, C.~Liu, H.~Zhao, S.~Pradhan, Y.~Chai, B.~Sapp, C.~Qi, Y.~Zhou, Z.~Yang, A.~Chouard, P.~Sun, J.~Ngiam, V.~Vasudevan, A.~Mccauley, J.~Shlens, D.~Anguelov, W.~Llc, and G.~Brain, ``Large scale interactive motion forecasting for autonomous driving: The waymo open motion dataset,'' in \emph{{ICCV}}, 2021.

\bibitem{Nayak2022-xw}
A.~Nayak, A.~Eskandarian, and Z.~Doerzaph, ``Uncertainty estimation of pedestrian future trajectory using bayesian approximation,'' \emph{IEEE Open Journal of Intelligent Transportation Systems}, vol.~3, pp. 617--630, 2022.

\bibitem{Richter2017-ca}
C.~Richter and N.~Roy, \emph{Safe Visual Navigation via Deep Learning and Novelty Detection}.\hskip 1em plus 0.5em minus 0.4em\relax Robotics: Science and Systems Foundation, Jul. 2017.

\bibitem{Fisac2018-yx}
J.~F. Fisac, A.~Bajcsy, S.~L. Herbert, D.~Fridovich-Keil, S.~Wang, C.~J. Tomlin, and A.~D. Dragan, ``Probabilistically safe robot planning with {Confidence-Based} human predictions,'' in \emph{RSS}, May 2018.

\bibitem{Ward2021-va}
T.~M. Ward, D.~M. Fer, Y.~Ban, G.~Rosman, O.~R. Meireles, and D.~A. Hashimoto, ``\BIBforeignlanguage{en}{Challenges in surgical video annotation},'' \emph{\BIBforeignlanguage{en}{Comput Assist Surg (Abingdon)}}, vol.~26, no.~1, pp. 58--68, Dec. 2021.

\bibitem{petermeijer2017driver}
S.~Petermeijer, F.~Doubek, and J.~De~Winter, ``Driver response times to auditory, visual, and tactile take-over requests: A simulator study with 101 participants,'' in \emph{2017 IEEE international conference on systems, man, and cybernetics (SMC)}.\hskip 1em plus 0.5em minus 0.4em\relax IEEE, 2017, pp. 1505--1510.

\bibitem{mehrotra2022human}
S.~Mehrotra, M.~Wang, N.~Wong, J.~Parker, S.~C. Roberts, W.~Kim, A.~Romo, and W.~J. Horrey, ``Human-machine interfaces and vehicle automation: A review of the literature,'' \emph{Accident Analysis \& Prevention}, vol. 109, pp. 18--28, 2022.

\bibitem{thorpe1996speed}
S.~Thorpe, D.~Fize, and C.~Marlot, ``Speed of processing in the human visual system,'' \emph{nature}, vol. 381, no. 6582, pp. 520--522, 1996.

\end{thebibliography}


\end{document}